\PassOptionsToPackage{table}{xcolor}
\documentclass{article} % For LaTeX2e
\usepackage{iclr2027_conference,times}
\usepackage[pagebackref=true,breaklinks=true,letterpaper=true,colorlinks,bookmarks=false]{hyperref}
\usepackage{url}

\usepackage[T1]{fontenc}
\usepackage{amsmath,amssymb,booktabs,array,tabularx,multirow}
\usepackage{graphicx,xcolor,microtype,placeins}
\usepackage{subcaption}
\usepackage{algorithm,algpseudocode}
\usepackage{wrapfig}
\usepackage{enumitem}
\newskip\recapOrigTextfloatsep \recapOrigTextfloatsep=\textfloatsep
\makeatletter
\let\recapOrigSection\section
\let\recapOrigSubsection\subsection
\def\section{\@startsection{section}{1}{\z@}{-1.6ex plus -0.4ex minus -.2ex}%
  {1.1ex plus 0.2ex minus 0.2ex}{\large\sc\raggedright}}
\def\subsection{\@startsection{subsection}{2}{\z@}{-1.4ex plus -0.4ex minus -.2ex}%
  {0.6ex plus .2ex}{\normalsize\sc\raggedright}}
\makeatother
\newcolumntype{Y}{>{\centering\arraybackslash}X}
\definecolor{oursrow}{RGB}{229,242,251}
\definecolor{denserow}{RGB}{242,242,242}
\definecolor{densered}{RGB}{170,35,35}
\definecolor{oursgreen}{RGB}{0,125,45}
\definecolor{boxgray}{gray}{0.6}
\definecolor{citeblue}{RGB}{20,60,150}
\definecolor{refred}{RGB}{215,15,15}
\hypersetup{citecolor=citeblue,urlcolor=black,linkcolor=refred}
\newcommand{\venue}[1]{\,{\normalfont\ttfamily\fontsize{6}{7}\selectfont\color{gray}(#1)}}
\newcommand{\oursbest}[1]{{\bfseries #1}}
\newcommand{\otherbest}[1]{{\bfseries #1}}
\newcommand{\secondbest}[1]{\underline{#1}}
\newcommand{\na}{--}
\newcommand{\method}{\textsc{Recap}}
\newcommand{\methodS}{\mbox{\textsc{Recap-speed}}}
\newcommand{\methodQ}{\mbox{\textsc{Recap-quality}}}

\newcommand{\figref}[1]{Fig.~\ref{#1}}
\newcommand{\figpanelref}[1]{Fig.~\ref{#1}}
\newcommand{\secref}[1]{Sec.~\ref{#1}}

\newcommand{\tabref}[1]{Tab.~\ref{#1}}
\newcommand{\eqnref}[1]{Eq.~\eqref{#1}}
\newcommand{\Figpanelref}[1]{Figure~\ref{#1}}
\newcommand{\Tabref}[1]{Table~\ref{#1}}
\newcommand{\Secref}[1]{Section~\ref{#1}}

\title{RECAP: Reciprocal Sparse Attention\\ and Hierarchical Caching\\ for Efficient Video Diffusion}

\author{Tiange Xu$^{1}$\thanks{Equal contribution.\protect\endgraf\noindent\hbox to1.8em{\hss$^{\dagger}$}Corresponding author: Yulun Zhang, yulun100@gmail.com} \enspace
Kai Liu$^{2*}$ \enspace Linghe Kong$^{2}$ \enspace Yulun Zhang$^{2\dagger}$ \\
$^{1}$University of Cambridge, $^{2}$Shanghai Jiao Tong University
}
\iclrfinalcopy % Uncomment for camera-ready version, but NOT for submission.

\begin{document}
\maketitle
\fancyhead[L]{Preprint}
\suppressfloats[t]
\vspace{-19pt}
\noindent\begin{minipage}{\linewidth}
  \centering
  \captionsetup{skip=4pt}
  \includegraphics[width=\linewidth]{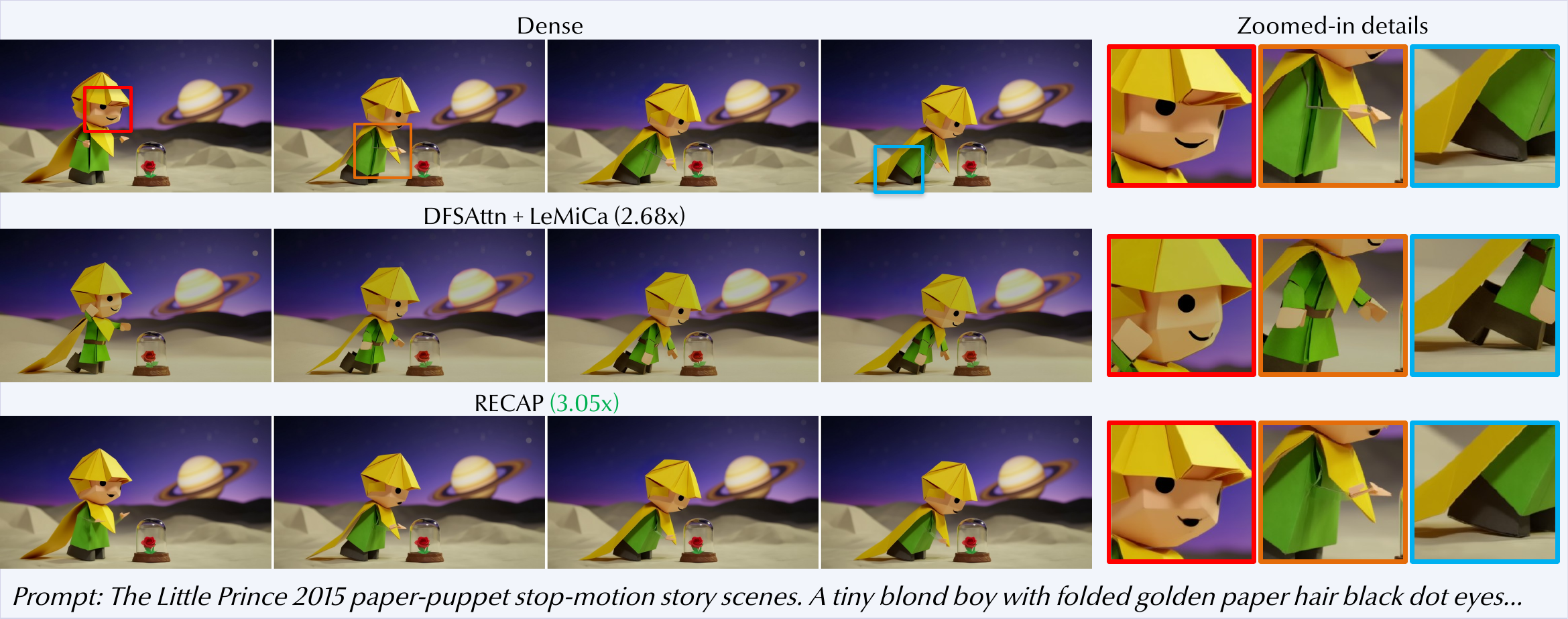}\par
  \captionof{figure}{Sample video and results on HunyuanVideo (720p, 5s).
\method{} achieves a 3.05$\times$ speedup while outperforming all tested
direct sparse--cache compositions in both fidelity and speed.
}
  \label{fig:teaser}
\end{minipage}
\vspace{1pt}

\begin{abstract}
\vspace{-3pt}
Diffusion transformers generate high-quality videos but remain
computationally expensive. Sparse attention and temporal caching reduce
computation within and across denoising steps, respectively, offering a
natural composition. However, predictions produced with sparse attention
become sources for temporal reuse, coupling their errors along the
sampling trajectory. We observe that improving the accuracy of these
sources recovers fidelity without changing the reuse schedule, and that
sparsification errors remain directionally aligned across nearby steps,
suggesting that correction information can be shared rather than
recomputed. These observations motivate \method{}, a training-free
method that couples sparse attention with hierarchical caching through
reciprocal correction, adaptively coordinated by online error
feedback. Cached dense refinement corrects omitted interactions in
current sparse estimates, while fresh sparse statistics track input
changes and regulate the aging correction. The resulting predictions
then renew the sources for whole-step prediction reuse, allowing prediction
history to advance between dense refreshes. We evaluate \method{} on
multiple video diffusion models against sparse-attention methods, caching
methods, and their direct compositions. On HunyuanVideo at 720p 5-second generation, \method{} achieves a
$3.05\times$ speedup and outperforms all tested direct compositions in
both fidelity and inference speed.
Code is available at \url{https://github.com/lyraxtg/RECAP}.

\end{abstract}

\vspace{-15pt}
\section{Introduction}
\label{sec:introduction}

Diffusion transformers have emerged as a powerful architecture for visual
generation~\citep{peebles2023scalable}, but the quadratic complexity of
attention and repeated transformer evaluations make inference expensive.
Video generation is especially costly due to the length of its
spatiotemporal token sequences~\citep{yang2025cogvideox,kong2024hunyuanvideo}. Sparse attention reduces
the interactions computed within each denoising step~\citep{xi2025sparse,hu2026dfsattn}, while temporal caching reuses features
or predictions across steps~\citep{liu2025teacache}. Since the two
approaches save computation along different axes, a natural way to
accelerate further is to compose them directly. In such a composition,
however, sparse-attention predictions become the sources that caching
reuses in later steps. Sparsification error therefore affects both the
current sampler update and subsequent reused predictions, making the
accuracy of these sources central to the speed--fidelity trade-off.

\begin{figure}[t]
  \centering
  \includegraphics[width=0.8\linewidth]{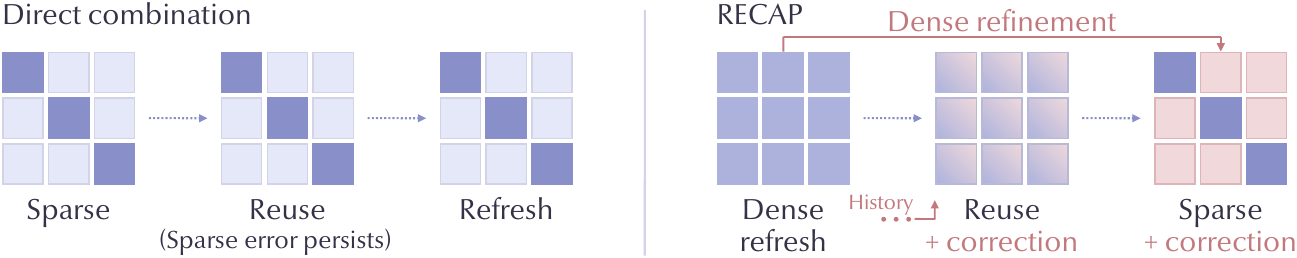}
  \captionsetup{skip=4pt}
  \caption{Motivation of \method{}. Direct composition reuses sparse outputs, so sparse error persists.
\method{} caches dense refinement to correct sparse steps, whose outputs renew the reuse history.}
  \label{fig:motivation}
\end{figure}

We study this coupling in direct compositions and make two observations.
First, source accuracy matters for fidelity. When we replace the cached
tensors with dense counterparts while keeping the reuse schedule and the
sparse predictions used by the sampler, the deviation from the dense
trajectory decreases. Computing every source densely, however, would
forfeit the savings of sparse attention, so we ask whether correction
can be shared across steps. Second, we find that sparsification errors
remain strongly aligned across nearby denoising steps. This persistence
suggests that a single dense correction can serve several sparse
evaluations, whose corrected predictions then provide more accurate
sources for temporal reuse. Sparse attention and caching should therefore
correct each other rather than act independently, as
\figref{fig:motivation} illustrates. \Secref{sec:motivation} details both analyses.

\begin{figure}[t]
  \centering
  % Cropped from archive/figures_old/local-method-figures/method_overview_redraw.pptx.
  \includegraphics[width=\linewidth]{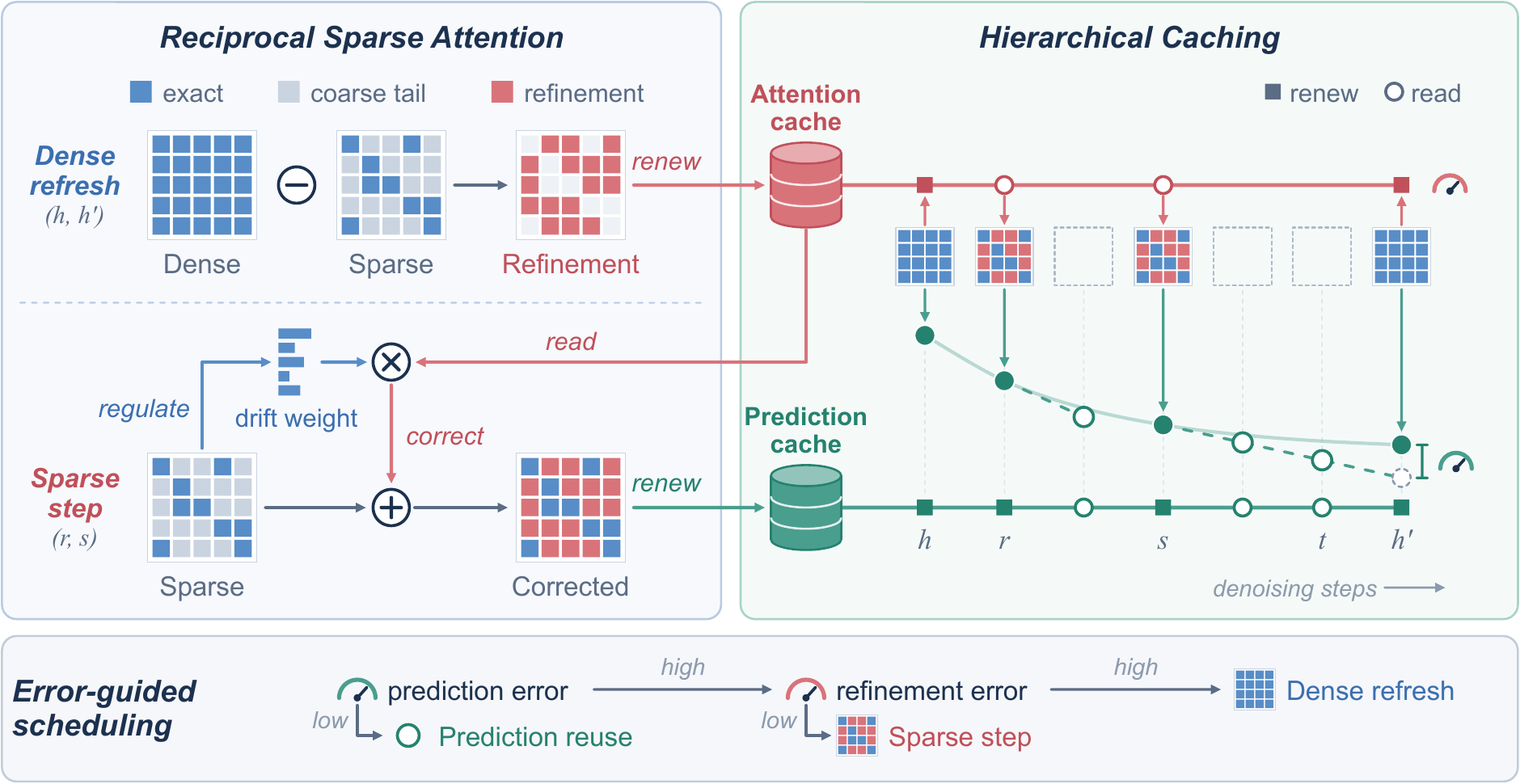}
  \captionsetup{skip=4pt}
  \caption{Overview of \method{}. A dense refresh caches the refinement that sparse attention misses, and sparse steps apply it with a drift weight from fresh statistics. Corrected outputs renew the prediction cache used for reuse, and the two errors choose among reuse, a sparse step, and a dense refresh.}
  \label{fig:overview}
\end{figure}

Motivated by these observations, we propose \method{}, which connects
sparse attention and a two-level cache hierarchy through reciprocal
correction, as illustrated in \figref{fig:overview}. At the attention level, a
cached dense refinement supplies the detail that the current sparse
estimate leaves unresolved. As attention inputs evolve across steps,
fresh sparse statistics in turn regulate how strongly this aging
refinement is applied, so that cached and current information correct
each other. At the prediction level, the corrected outputs renew a
prediction cache, from which later steps are extrapolated without another
transformer evaluation. Sparse computation thus links the two levels,
and one dense refresh can support several sparse evaluations together
with the prediction reuse that follows them.

This design also naturally supplies the feedback for adaptive online scheduling. Each computed prediction reveals the error of the extrapolation it replaces, and each dense refresh reveals the error of the refinement it renews. Since corrected sparse evaluations renew prediction history while reusing existing refinement, the two caches can be renewed at different rates. The prediction error thus decides when to compute a new prediction, and the refinement error decides whether this computation needs a dense refresh. Under a fixed budget, these decisions choose among prediction reuse, corrected sparse computation, and dense refresh, adapting the schedule to each sample without offline search.

Our contributions are summarized as follows:
\begin{itemize}[leftmargin=2em,topsep=2pt,itemsep=1pt,parsep=0pt,beginpenalty=10000]
  \item We identify source accuracy and temporal persistence of
  sparsification error as key opportunities to improve sparse--cache
  composition. More accurate cache sources recover fidelity, while
  aligned errors across steps motivate sharing dense correction
  across sparse evaluations.

  \item We propose \method{}, which couples sparse attention with
  hierarchical caching through reciprocal correction. Cached dense refinement
  corrects sparse attention, corrected predictions renew the sources
  for whole-step reuse, and online error feedback decides when each cache is renewed.

  \item We evaluate \method{} on three video diffusion models, where it
  improves speed--fidelity trade-offs over individual accelerators and
  direct sparse--cache compositions. On HunyuanVideo, it achieves a
  3.05$\times$ speedup with higher fidelity than all tested compositions, as \figref{fig:teaser} shows.
\end{itemize}

\section{Related Work}
\label{sec:related-work}

\textbf{Sparse attention.}
Sparse attention exploits spatial and temporal patterns~\citep{xi2025sparse,zhang2025fast,li2025radial}, or selects important
interactions through content-dependent scores~\citep{xia2025training,zhang2025spargeattn},
token reordering~\citep{yang2025sparse2,hu2026dfsattn}, and query--key
clustering~\citep{luo2026attention}. Training-based approaches further
learn sparse patterns~\citep{zhang2025vsa,zhang2026sla}. Instead of
discarding omitted interactions, PISA approximates them with block means
and shared first-order value compensation~\citep{li2026pisa}, and COBS
estimates block mass with second-order key statistics~\citep{tian2026cobs}.
Building on these approximations, \method{} corrects the remaining tail
value and normalization mass with dense refinement cached from earlier steps.

\textbf{Temporal caching.}
Caching exploits similarity across denoising steps by reusing features or
outputs~\citep{ma2024deepcache,chen2024delta,zhao2025pab}.
LeMiCa optimizes reuse offline~\citep{gao2025lemica}, while
other methods adapt reuse to online signals~\citep{liu2025teacache,kahatapitiya2024adacache,lv2026navicache,ma2025magcache,zhou2025easycache}.
Further approaches include token-level reuse~\citep{zou2025toca}, feature extrapolation~\citep{liu2025taylorseer,liu2026d2cache}, guidance sharing~\citep{lv2025fastercache},
and learned routing~\citep{ma2024learningtocache}.
\method{} uses current sparse statistics to regulate aging refinement
and corrected sparse predictions to renew the sources for whole-step prediction reuse.
Rather than deciding only when to reuse, \method{} also improves the predictions
that are reused.

\textbf{Combining sparse attention and temporal reuse.}
DiTFastAttn shares dense-minus-window attention residuals and attention outputs
under an offline-selected schedule~\citep{yuan2024ditfastattn}.
Re-ttention reuses normalization ratios and output residuals to reconstruct
sparse attention~\citep{chen2025rettention}.
ASTRAEA uses evolutionary search to optimize per-timestep token budgets~\citep{liu2026astraea}, while LoSA reuses early-step attention block
indices across subsequent denoising steps~\citep{liu2026losa}.
These methods act in one direction only, whereas \method{} corrects
reciprocally and coordinates both cache levels by online error feedback.
Appendix Sec.~F discusses these distinctions further.

\section{Method}
\label{sec:method}
\vspace{-4pt}
\subsection{Motivation}
\label{sec:motivation}

\textbf{Background.}
At denoising step $t$, a diffusion transformer predicts
$p_t=f_\theta(x_t,t,c)$ from latent $x_t$ and conditioning $c$,
and the sampler advances as $x_{t+1}=\Psi_t(x_t,p_t)$.
Sparse attention reduces computation within each evaluation by omitting
or approximating query-key interactions, while caching reuses
earlier features or predictions across steps.
These complementary savings make their combination a natural way to
further accelerate sampling.
However, sparse outputs also become sources for subsequent reuse,
so sparsification error can affect both the current sampler update
and later steps.
This coupling motivates a closer examination of how the two mechanisms
interact.

\begin{figure}[t]
  \centering
  \includegraphics[width=\linewidth]{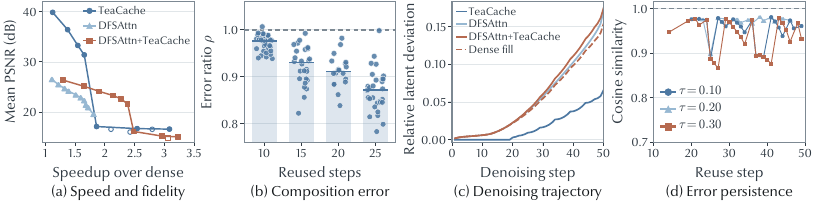}
  \captionsetup{skip=4pt}
  \caption{
(a) Speed and fidelity of direct composition.
(b) Composed error norm relative to the norm of summed standalone errors.
(c) Latent deviation with dense cache filling under a fixed reuse schedule.
(d) Cosine similarity of sparsification errors between source and reuse
steps for different TeaCache thresholds $\tau$. Details of these experiments are provided in Appendix Sec.~C.1.}
  \label{fig:evidence}
  \vspace{-\parskip}\nointerlineskip
  \begin{subfigure}{0pt}
  \phantomsubcaption\label{fig:evidence-a}
  \phantomsubcaption\label{fig:evidence-b}
  \phantomsubcaption\label{fig:evidence-c}
  \phantomsubcaption\label{fig:evidence-d}
  \end{subfigure}
\end{figure}

\textbf{Composition error and source accuracy.}
We sweep the settings of DFSAttn, TeaCache, and their direct composition
in \figpanelref{fig:evidence-a}, finding that combining the two
approximations need not cause a disproportionate loss of fidelity.
If the two errors simply accumulated, the composed error would match the vector
sum of the standalone errors. Instead, it falls below this sum in 103 of 104
configurations in \figpanelref{fig:evidence-b}, revealing a favorable interaction
between sparse attention and caching.

To reduce the remaining error, we focus on source accuracy, since sparse
evaluations supply the sources of subsequent reuse.
In DFSAttn with TeaCache, we replace cached tensors with dense
counterparts computed on the same inputs, while retaining the sparse
predictions for sampler updates and the reuse schedule.
As \figpanelref{fig:evidence-c} shows, this reduces latent
deviation from the matched dense trajectory, demonstrating that \emph{more
accurate cache sources can improve the fidelity of the composition.}

\textbf{Persistence of sparsification error.}
Computing every source densely, however, would forfeit the savings of sparse attention. We therefore ask whether correction can instead be shared across steps. Along dense
trajectories, we compare sparse and dense predictions on identical inputs
to isolate sparsification error from trajectory drift.
\Figpanelref{fig:evidence-d} shows that these errors remain strongly
aligned across nearby steps for all thresholds, so a correction measured at
the source step still points in the right direction at later steps. This
suggests that \emph{a single dense correction can support multiple sparse
evaluations} by combining a fresh sparse estimate with cached detail that is
costly to recompute.

\textbf{Method overview.}
\method{} connects these insights through a two-level cache hierarchy, as
illustrated in \figref{fig:overview}. The persistence of sparsification
error motivates sharing dense correction across sparse evaluations,
which we develop as attention-level refinement in
\secref{sec:reconstruction}. The resulting improved predictions then
serve as more accurate sources for whole-step temporal reuse in \secref{sec:prediction-reuse}. Finally, \secref{sec:scheduling} coordinates
renewal at both levels through online error feedback. At each step, \method{}
either reuses an extrapolated prediction, performs corrected sparse
computation, or refreshes densely.

\subsection{Attention-Level Caching for Sparse Reconstruction}
\label{sec:reconstruction}

The source-accuracy findings in \secref{sec:motivation} motivate
improving sparse predictions before reuse, and the temporal alignment
of sparsification error suggests sharing correction across steps.
Since this error originates in the omitted interactions,
we realize both ideas at the attention level, where a fresh
coarse estimate tracks changes in omitted attention and cached dense
refinement supplies the discrepancy that this estimate leaves unresolved.
Both are maintained separately for each layer and attention head.

\textbf{Selected interactions and coarse tail.}
For scaled dot-product attention~\citep{vaswani2017attention}, we partition
the keys for a query $q$ into a selected set $H$ and a tail $T$ omitted
from exact computation.
We select key blocks by pooled query-key scores and compute attention
over $H$ exactly.
The omitted tail contributes both attention values and normalization
mass, so reconstructing the full output requires estimating both.
For either set $A$, let $o_A$ be its normalized attention output and
$\ell_A=\log\sum_{j\in A}\exp(\beta q^\top k_j)$ its log-mass,
where $\beta=d_k^{-1/2}$.
The full output is
\begin{equation}
  o_t=\frac{e^{\ell_{H,t}}o_{H,t}+e^{\ell_{T,t}}o_{T,t}}
                 {e^{\ell_{H,t}}+e^{\ell_{T,t}}}.
  \label{eq:merge}
\end{equation}

For the omitted tail, block means provide initial estimates of value
and normalization mass, which we refine through two moment corrections.
Key variance refines the log-mass through the second-order approximation
of COBS~\citep{tian2026cobs}, while pooled value-key covariance refines
the value through the first-order compensation of PISA~\citep{li2026pisa}.
The logarithm of the summed block masses gives the coarse log-mass
$\bar\ell_{T,t}$, while weighting block values by their normalized masses
gives the coarse output $\bar o_{T,t}$.
These statistics are recomputed at each sparse evaluation, allowing the
estimate to track attention changes without evaluating every omitted
query-key interaction.

\textbf{Cached dense refinement.}
The coarse estimate leaves a discrepancy that a dense refresh can recover.
At refresh $h$, we subtract the selected branch's mass and unnormalized
value contribution from those of full attention to obtain the exact tail
output $o_{T,h}$ and log-mass $\ell_{T,h}$.
We cache their differences from the coarse estimates and add these
residuals to fresh estimates at a later sparse evaluation $t$,
\begin{equation}
\begin{aligned}
  d_h^v&=o_{T,h}-\bar o_{T,h},
  &d_h^\ell&=\ell_{T,h}-\bar\ell_{T,h},\\
  \widehat o_{T,t}&=\bar o_{T,t}+d_h^v,
  &\widehat\ell_{T,t}&=\bar\ell_{T,t}+d_h^\ell.
\end{aligned}
\label{eq:refinement}
\end{equation}
Each layer retains its selected partition between dense refreshes, so
that cached residuals and the current coarse estimates continue to
describe the same set of omitted keys. The partition can then be re-selected at each
dense refresh, together with the cached residuals, as attention evolves.

\begin{figure}[!t]
  \centering
  \includegraphics[width=\linewidth]{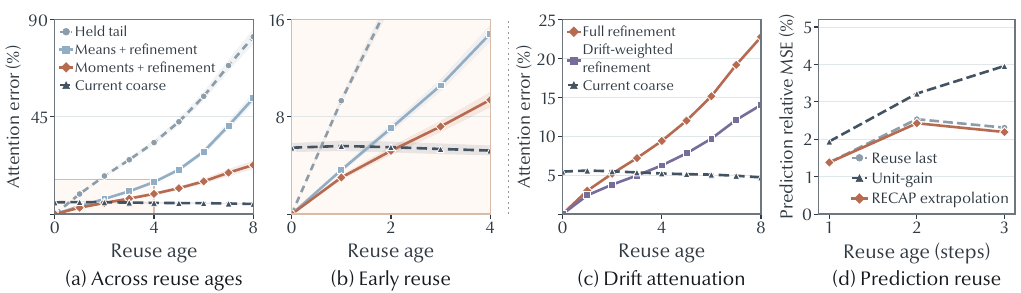}
  \captionsetup{skip=4pt}
  \caption{Attention refinement and prediction reuse.
  (a, b) Reconstruction error growth with reuse age.
  (c) Drift weighting moderates error growth relative to full-strength
  refinement.
  (d) Fitted gain extrapolation compared with holding the latest
  prediction and unit-gain extrapolation.}
  \label{fig:refinement}
  \vspace{-\parskip}\nointerlineskip
  \begin{subfigure}{0pt}
    \phantomsubcaption\label{fig:refinement-a}
    \phantomsubcaption\label{fig:refinement-b}
    \phantomsubcaption\label{fig:refinement-c}
    \phantomsubcaption\label{fig:refinement-d}
  \end{subfigure}
\end{figure}

To see why this decomposition helps, write either tail quantity as
$y_t=\bar y_t+d_t$, so that the reconstruction error becomes
$\widehat y_t-y_t=d_h-d_t$. The fresh estimate absorbs changes in the
coarse component, leaving reconstruction error determined only by
changes in the unresolved discrepancy. As \figpanelref{fig:refinement-a}
and \figpanelref{fig:refinement-b} illustrate, updating the coarse
component slows error growth relative to holding the entire tail fixed, while
adding cached refinement further improves accuracy at early reuse ages.

\textbf{Drift-weighted refinement.}
However, \figpanelref{fig:refinement-b} also shows that the benefit of
cached refinement diminishes with reuse age as attention inputs evolve.
We therefore use the fresh coarse statistics to regulate the cached
contribution, so that current sparse computation in turn corrects the cache
when earlier refinement becomes less reliable.
To quantify this drift, each dense refresh also stores the coarse
log-mass $\bar\ell_{T,i,h}$ alongside the residuals. For video query
row $i$, the change $u_{i,t}=\bar\ell_{T,i,t}-\bar\ell_{T,i,h}$
captures how far current tail statistics have moved from those at the
refresh. We center this over the $N_q$ video query rows within each
sample and attention head to obtain the refinement weight
\begin{equation}
  \Delta_{i,t}=u_{i,t}-\frac{1}{N_q}\sum_{j=1}^{N_q}u_{j,t},
  \qquad a_{i,t}=e^{-|\Delta_{i,t}|}.
  \label{eq:drift}
\end{equation}
Centering removes the common log-mass shift already carried by the
fresh estimate. The weight thus decreases only with row-specific
deviations and introduces no tunable parameter.
To keep value and mass consistent, the same weight blends their
unnormalized contributions,
\begin{equation}
\begin{aligned}
  \widetilde Z_{T,t}
    &=(1-a_t)e^{\bar\ell_{T,t}}+a_t e^{\widehat\ell_{T,t}},\\
  \widetilde N_{T,t}
    &=(1-a_t)e^{\bar\ell_{T,t}}\bar o_{T,t}
       +a_t e^{\widehat\ell_{T,t}}\widehat o_{T,t}.
\end{aligned}
  \label{eq:attenuation}
\end{equation}
The normalized output
$\widetilde o_{T,t}=\widetilde N_{T,t}/\widetilde Z_{T,t}$
and log-mass $\widetilde\ell_{T,t}=\log\widetilde Z_{T,t}$
then merge with the selected branch through \eqnref{eq:merge}.
As shown in \figpanelref{fig:refinement-c}, this weighting moderates
reconstruction error growth compared with applying refinement at
full strength as the cached correction ages.
Appendix Sec.~A provides the mathematical details of
the moment estimates, tail recovery, and blending.

\subsection{Prediction-Level Caching for Step Reuse}
\label{sec:prediction-reuse}

Since more accurate sources improve the fidelity of temporal reuse,
the corrected sparse predictions from \secref{sec:reconstruction}
naturally serve as better anchors. This motivates caching and
extrapolating these improved outputs to advance subsequent sampler
steps without another transformer evaluation.

\textbf{Prediction extrapolation.}
To account for changes between steps, we estimate a local rate of
change from the two most recent computed predictions at steps $r<s$
and extrapolate to a later step $t$,
\begin{equation}
  v_s=\frac{p_s-p_r}{s-r},
  \qquad \widehat p_t=p_s+\alpha_s(t-s)v_s,
  \label{eq:prediction}
\end{equation}
where intervals count sampler steps and the scalar gain $\alpha_s$
controls how strongly the observed change is carried forward.
Because prediction changes may not continue at the same rate, we fit
the gain online. At each computed step $t$ with two preceding
sources, we compare the observed change $y_t=p_t-p_s$ with the
unit-gain extrapolation $z_t=(t-s)v_s$ through a cumulative
least-squares fit,
\begin{equation}
  \alpha_t=\arg\min_{a\in\mathbb R}
  \sum_{j\in\mathcal C_t}\|y_j-a z_j\|_F^2,
  \label{eq:gain-fit}
\end{equation}
where $\mathcal C_t$ contains computed steps through $t$ with two
preceding sources. During prediction
reuse, the sampler advances with $\widehat p_t$ while the transformer
evaluation is skipped and both the prediction history and gain
remain fixed. As \figpanelref{fig:refinement-d} illustrates, the fitted gain
reduces extrapolation error relative to unit gain and improves over
simply holding the latest prediction at longer reuse ages. This comparison fixes
prediction histories and latent inputs to isolate the effect of the extrapolation rule.

\textbf{Prediction-source renewal.}
Each computed prediction updates both the anchor and the rate of
change, so source accuracy directly affects both terms in
\eqnref{eq:prediction}. Corrected sparse evaluations renew these
quantities while continuing to reuse existing attention refinement,
allowing prediction history to advance between dense refreshes
without requiring a new dense correction. A single dense refresh
can therefore support several sparse evaluations and their subsequent
prediction reuse. Restricting this renewal to
dense refreshes substantially reduces fidelity, as \secref{sec:ablations} shows. Appendix Sec.~B.1 derives the closed-form
gain and shows how source errors are carried into reused predictions.

\subsection{Error-Guided Online Scheduling}
\label{sec:scheduling}

The two levels of reuse in \method{} naturally supply the feedback
needed to schedule them. Each computed prediction reveals the error of
the extrapolation it replaces, and each dense refresh reveals the error
of the refinement it renews. \method{} uses these errors to decide online
when to compute a new prediction and whether this computation needs a
dense refresh. The schedule therefore adapts to each sample as sampling
proceeds, rather than relying on offline search or a fixed reuse pattern.

\textbf{Prediction feedback decides when to compute.}
We compare each computed prediction $p_t$ with its extrapolation
$\widehat p_t$, normalizing by the elapsed steps since extrapolation
error grows with the interval,
\begin{equation}
  e_{\mathrm{pred},t}
    =\frac{\|p_t-\widehat p_t\|_F}{\|p_t\|_F},
  \qquad \rho_{\mathrm{pred},t}
    =\frac{e_{\mathrm{pred},t}}{t-s},
  \label{eq:prediction-check}
\end{equation}
where $s$ is the latest source step. A rate below $\tau_{\mathrm{pred}}$
extends the next reuse interval, which is capped by spreading the
remaining evaluations of a budget $B$ evenly over the remaining steps.
This cap shortens when the rate exceeds its running mean, which
concentrates computation where predictions change quickly. The budget
thus fixes how much is computed, while the feedback decides where.

\textbf{Refinement feedback decides how to compute.}
Corrected sparse attention is used only while the cached refinement
remains reliable, which requires a dense reference to verify. At each
dense refresh~$h$, we compare the dense attention output $O_{l,h}^D$ of
each layer $l$ with the corrected sparse output
$\widetilde O_{l,h}^{\mathrm{old}}$ reconstructed from the old refinement.
Pooling the layer errors before the residual update gives
\begin{equation}
  e_{\mathrm{ref},h}=
  \left(
    \frac{\sum_l\|\widetilde O_{l,h}^{\mathrm{old}}-O_{l,h}^D\|_F^2}
         {\sum_l\|O_{l,h}^D\|_F^2}
  \right)^{1/2}.
  \label{eq:refinement-check}
\end{equation}
Since sparsification error remains aligned across nearby steps, as
\figpanelref{fig:evidence-d} shows, this error indicates how long the
new refinement can be trusted. A passing check extends its permitted
age, whereas a failed check makes the next computed prediction dense.
Between checks, drift weighting adjusts the cached contribution per
query row, while the check decides when to renew the whole refinement.

Together, the two decisions yield the three actions in
\figref{fig:overview}, namely prediction reuse, corrected sparse
computation that renews only the prediction level, and a dense refresh
that renews both cache levels. Each computed step thus supplies both a new source for
reuse and the feedback that decides when the two caches are renewed.
Appendix Sec.~B gives the update
rules and sampling algorithm.

\begin{table}[t]
  \caption{Main results on CogVideoX-1.5 and HunyuanVideo. Re-ttention exceeds GPU memory on HunyuanVideo and is therefore omitted. VBench Overall averages the six evaluated dimensions after per-dimension normalization.
Best and second-best results are bold and underlined, respectively.}
  \label{tab:main-results-single}
  \centering
  \setlength{\tabcolsep}{0.8pt}
  \renewcommand{\arraystretch}{1.00}
  \fontsize{7.5}{8.5}\selectfont
  \begin{tabularx}{\textwidth}{@{}l|*{3}{Y}|*{5}{Y}|*{2}{Y}@{}}
    \toprule
    \multicolumn{1}{c|}{\multirow{2}{*}{\textbf{Method}}} &
    \multicolumn{3}{c|}{\textbf{Fidelity}} &
    \multicolumn{5}{c|}{\textbf{Quality}} &
    \multicolumn{2}{c}{\textbf{Efficiency}} \\
    \cmidrule(lr){2-4}\cmidrule(lr){5-9}\cmidrule(l){10-11}
    & PSNR$\uparrow$ & SSIM$\uparrow$ & LPIPS$\downarrow$ &
    {\fontsize{5.5}{7}\selectfont CLIPIQA+}$\uparrow$ & IQ$\uparrow$ & AQ$\uparrow$ & TF$\uparrow$ & Overall$\uparrow$ &
    {\fontsize{7}{8}\selectfont Latency}$\downarrow$ &
    {\fontsize{6.5}{8}\selectfont Speedup}$\uparrow$ \\
    \midrule
    \multicolumn{11}{c}{\itshape CogVideoX-1.5 (768$\times$1360, 81 frames)} \\
    \midrule
    \rowcolor{denserow}
    Dense & \na & \na & \na & 0.489 & 64.05 & 61.33 & 97.06 & 83.31 & 1,106s & 1.00$\times$ \\
    TeaCache\venue{CVPR'25} & 31.13 & 0.9472 & 0.0475 & 0.472 & 63.16 & 60.95 & 97.06 & 83.06 & 664s & 1.67$\times$ \\
    SVG\venue{ICML'25} & 23.17 & 0.8044 & 0.1952 & 0.478 & 63.41 & 60.16 & 96.76 & 82.69 & 776s & 1.43$\times$ \\
    Re-ttention\venue{NeurIPS'25} & 22.42 & 0.7587 & 0.2557 & 0.373 & 57.67 & 56.38 & 96.98 & 81.16 & 837s & 1.32$\times$ \\
    LeMiCa\venue{NeurIPS'25} & 32.06 & 0.9631 & 0.0359 & 0.480 & 63.31 & 61.05 & 97.07 & 83.09 & 700s & 1.58$\times$ \\
    SVOO\venue{ICML'26} & 25.88 & 0.8532 & 0.1484 & 0.451 & 63.50 & 60.12 & 96.91 & 82.75 & 778s & 1.42$\times$ \\
    DFSAttn\venue{ICML'26} & 26.13 & 0.8694 & 0.1266 & 0.480 & 63.32 & 60.58 & 96.88 & 82.85 & 685s & 1.61$\times$ \\
    NaviCache\venue{ICML'26} & 26.99 & 0.8788 & 0.1123 & 0.468 & 63.22 & 60.68 & \otherbest{97.10} & 83.04 & \secondbest{654s} & \secondbest{1.69$\times$} \\
    \rowcolor{oursrow}
    \textbf{\methodS{}} & \secondbest{33.69} & \secondbest{0.9675} & \secondbest{0.0306} & \secondbest{0.483} & \secondbest{63.56} & \secondbest{61.08} & \secondbest{97.09} & \secondbest{83.16} & \oursbest{632s} & \oursbest{1.75$\times$} \\
    \rowcolor{oursrow}
    \textbf{\methodQ{}} & \oursbest{35.80} & \oursbest{0.9766} & \oursbest{0.0223} & \oursbest{0.486} & \oursbest{63.79} & \oursbest{61.28} & 97.06 & \oursbest{83.23} & 681s & 1.62$\times$ \\
    \midrule
    \multicolumn{11}{c}{\itshape HunyuanVideo (720$\times$1280, 129 frames)} \\
    \midrule
    \rowcolor{denserow}
    Dense & \na & \na & \na & 0.512 & 64.91 & 58.74 & 98.92 & 85.13 & 3,440s & 1.00$\times$ \\
    TeaCache\venue{CVPR'25} & 21.94 & 0.8376 & 0.2073 & 0.495 & 62.82 & 57.96 & 98.92 & 84.57 & \secondbest{1,638s} & \secondbest{2.10$\times$} \\
    SVG\venue{ICML'25} & 21.13 & 0.8203 & 0.2458 & 0.490 & 63.76 & 58.26 & 98.76 & 84.51 & 1,641s & 2.10$\times$ \\
    LeMiCa\venue{NeurIPS'25} & 35.12 & 0.9730 & 0.0410 & \otherbest{0.513} & 63.12 & 58.45 & \secondbest{99.03} & 84.87 & 1,746s & 1.97$\times$ \\
    SVOO\venue{ICML'26} & 27.68 & 0.9048 & 0.1218 & 0.505 & 64.11 & 58.40 & 98.91 & 84.85 & 1,831s & 1.88$\times$ \\
    DFSAttn\venue{ICML'26} & 30.67 & 0.9420 & 0.0624 & 0.504 & 64.14 & 58.45 & 98.80 & 84.78 & 1,697s & 2.03$\times$ \\
    NaviCache\venue{ICML'26} & 35.29 & 0.9720 & 0.0310 & 0.502 & \otherbest{64.63} & \secondbest{58.68} & 98.94 & 85.02 & 1,761s & 1.95$\times$ \\
    \rowcolor{oursrow}
    \textbf{\methodS{}} & \secondbest{35.35} & \secondbest{0.9744} & \secondbest{0.0268} & \secondbest{0.508} & 63.82 & 58.66 & \oursbest{99.07} & \secondbest{85.04} & \oursbest{1,611s} & \oursbest{2.14$\times$} \\
    \rowcolor{oursrow}
    \textbf{\methodQ{}} & \oursbest{36.71} & \oursbest{0.9795} & \oursbest{0.0189} & \oursbest{0.513} & \secondbest{64.27} & \oursbest{58.73} & 98.96 & \oursbest{85.08} & 1,727s & 1.99$\times$ \\
    \bottomrule
  \end{tabularx}
  \vspace{3pt}
\end{table}

\begin{table}[!t]
\caption{Comparison with direct sparse--cache compositions, each configured to a similar speedup in the high-speed regime. Best and second-best results are bold and underlined, respectively.}

  \label{tab:lower-band}
  \centering
  \setlength{\tabcolsep}{0.8pt}
  \renewcommand{\arraystretch}{1.00}
  \fontsize{7.5}{8.5}\selectfont
  \begin{tabularx}{\textwidth}{@{}>{\raggedright\arraybackslash}p{2.65cm}|*{3}{Y}|*{5}{Y}|*{2}{Y}@{}}
    \toprule
    \multicolumn{1}{c|}{\multirow{2}{*}{\textbf{Method}}} &
    \multicolumn{3}{c|}{\textbf{Fidelity}} &
    \multicolumn{5}{c|}{\textbf{Quality}} &
    \multicolumn{2}{c}{\textbf{Efficiency}} \\
    \cmidrule(lr){2-4}\cmidrule(lr){5-9}\cmidrule(l){10-11}
    & PSNR$\uparrow$ & SSIM$\uparrow$ & LPIPS$\downarrow$
    & {\fontsize{5.5}{7}\selectfont CLIPIQA+}$\uparrow$ & IQ$\uparrow$ & AQ$\uparrow$ & TF$\uparrow$ & Overall$\uparrow$
    & {\fontsize{7}{8}\selectfont Latency}$\downarrow$
    & {\fontsize{6.5}{8}\selectfont Speedup}$\uparrow$ \\
    \midrule

    \multicolumn{11}{c}{\itshape CogVideoX-1.5 (768$\times$1360, 81 frames)} \\
    \midrule
    DFSAttn + TeaCache
    & 15.89 & 0.5654 & 0.4245 & 0.422 & 61.59 & 58.32 & 95.93 & 81.55 & 446s & 2.48$\times$ \\
    DFSAttn + LeMiCa
    & \secondbest{21.57} & \secondbest{0.7835} & \secondbest{0.1934}
    & \secondbest{0.437} & \secondbest{63.43} & \secondbest{60.76} & 95.91 & 82.30 & 440s & 2.51$\times$ \\
    DFSAttn + NaviCache
    & 21.42 & 0.7648 & 0.2059 & 0.434 & 63.39 & 60.42 & 96.04 & 82.23 & 434s & 2.54$\times$ \\
    SVOO + TeaCache
    & 15.77 & 0.5594 & 0.4206 & 0.419 & 62.79 & 56.69 & 95.83 & 81.41 & 439s & 2.52$\times$ \\
    SVOO + LeMiCa
    & 21.14 & 0.7627 & 0.2043 & 0.430 & 63.34 & 60.34 & 96.07 & \secondbest{82.32} & 438s & 2.52$\times$ \\
    SVOO + NaviCache
    & 20.43 & 0.7047 & 0.2816 & 0.405 & 62.95 & 58.34 & \secondbest{96.17} & 81.72
    & \secondbest{427s} & \secondbest{2.59$\times$} \\
    \rowcolor{oursrow}
    \textbf{\method{}}
    & \oursbest{25.35} & \oursbest{0.8724} & \oursbest{0.1421}
    & \oursbest{0.439} & \oursbest{63.61} & \oursbest{60.86} & \oursbest{96.30} & \oursbest{82.64}
    & \oursbest{415s} & \oursbest{2.67$\times$} \\
    \midrule

\multicolumn{11}{c}{\itshape HunyuanVideo (720$\times$1280, 129 frames)} \\
\midrule
DFSAttn + TeaCache
& 24.30 & 0.8701 & 0.1545 & 0.464
& \otherbest{62.71} & 58.14 & 98.43 & 83.91
& \secondbest{1,213s} & \secondbest{2.83$\times$} \\

DFSAttn + LeMiCa
& \secondbest{28.49} & 0.9146 & 0.0905
& \secondbest{0.470} & 60.69 & 57.80 & 98.69 & 83.79
& 1,283s & 2.68$\times$ \\

DFSAttn + NaviCache
& 28.46 & \secondbest{0.9148} & \secondbest{0.0875}
& 0.464 & 62.18 & 58.14 & 98.48 & 83.80
& 1,350s & 2.55$\times$ \\

SVOO + TeaCache
& 21.27 & 0.8092 & 0.2535 & 0.461
& 60.95 & 57.85 & 98.45 & 83.52
& 1,447s & 2.38$\times$ \\

SVOO + LeMiCa
& 27.39 & 0.8999 & 0.1266
& \otherbest{0.471} & 61.18 & 57.97
& \secondbest{98.73} & 83.86
& 1,329s & 2.59$\times$ \\

SVOO + NaviCache
& 27.16 & 0.8976 & 0.1280 & 0.467
& 62.31 & \secondbest{58.35} & 98.64
& \secondbest{83.97}
& 1,560s & 2.21$\times$ \\

\rowcolor{oursrow}
\textbf{\method{}}
& \oursbest{29.67} & \oursbest{0.9392} & \oursbest{0.0731}
& 0.467 & \secondbest{62.35} & \oursbest{58.68}
& \oursbest{98.81} & \oursbest{84.34}
& \oursbest{1,129s} & \oursbest{3.05$\times$} \\
\bottomrule
  \end{tabularx}

  \par\vspace{4pt}
  \begin{minipage}{\linewidth}
  \centering\begingroup
  \setlength{\fboxsep}{0pt}\setlength{\fboxrule}{0.45pt}
  \newcommand{\stackcol}{0.164\linewidth}
  \newcommand{\stacklabel}[2]{\begin{minipage}[c]{\stackcol}\centering\fontsize{7.5}{9}\selectfont\textcolor{#2}{\mbox{#1}}\end{minipage}}
  \newcommand{\stackmetric}[3]{\begin{minipage}[c]{\stackcol}\centering\fontsize{7.5}{9}\selectfont\textcolor{#3}{#1$\times$ / #2}\end{minipage}}
  \newcommand{\stackdensemetric}{\begin{minipage}[c]{\stackcol}\centering\fontsize{7.5}{9}\selectfont\textcolor{black}{Speedup / PSNR}\end{minipage}}
  \newcommand{\stackboxed}[3]{\begin{minipage}[c]{\stackcol}\makebox[\linewidth][c]{\fcolorbox{#3}{white}{\includegraphics[width=\dimexpr\linewidth-2\fboxrule\relax]{figures/stacked_comparison/#1_#2.jpg}}}\end{minipage}}
  \ExplSyntaxOn
  \cs_set_protected:Npn \stackfitprompt #1
    {
      \group_begin:
      \fontsize{8.5}{10}\selectfont\itshape
      \seq_set_split:Nnn \l_tmpa_seq { ~ } {#1}
      \tl_set:Nn \l_tmpa_tl { Prompt: }
      \seq_map_inline:Nn \l_tmpa_seq
        {
          \hbox_set:Nn \l_tmpa_box { \tl_use:N \l_tmpa_tl ~ ##1 ... }
          \dim_compare:nNnTF { \box_wd:N \l_tmpa_box } > { \linewidth }
            { \tl_put_right:Nn \l_tmpa_tl { ... } \seq_map_break: }
            { \tl_put_right:Nn \l_tmpa_tl { ~ ##1 } }
        }
      \makebox[\linewidth][c]{\tl_use:N \l_tmpa_tl}\par
      \group_end:
    }
  \ExplSyntaxOff
  \newcommand{\stackrow}[1]{%
    \noindent\stackboxed{#1}{0}{densered}\hfill\stackboxed{#1}{1}{boxgray}\hfill\stackboxed{#1}{2}{boxgray}\hfill\stackboxed{#1}{3}{boxgray}\hfill\stackboxed{#1}{4}{boxgray}\hfill\stackboxed{#1}{5}{blue!75!black}\par}
  \noindent
  \stacklabel{\textbf{Dense}}{densered}\hfill%
  \stacklabel{DFSAttn + TeaCache}{black}\hfill%
  \stacklabel{DFSAttn + NaviCache}{black}\hfill%
  \stacklabel{SVOO + TeaCache}{black}\hfill%
  \stacklabel{SVOO + LeMiCa}{black}\hfill%
  \stacklabel{\textbf{\method{}}}{blue!75!black}\par\vspace{2pt}
  \stackrow{cog}
  \vspace{0.7pt}
  \noindent
  \stackdensemetric\hfill%
  \stackmetric{2.42}{16.71}{black}\hfill%
  \stackmetric{2.45}{19.93}{black}\hfill%
  \stackmetric{2.47}{17.50}{black}\hfill%
  \stackmetric{2.48}{17.83}{black}\hfill%
  \stackmetric{2.57}{21.89}{blue!75!black}\par
  \vspace{1pt}
  \stackfitprompt{A majestic eagle soars gracefully above a vast, snow-covered forest, its powerful wings cutting through the crisp winter air.}
  \vspace{2pt}
  \nointerlineskip\hbox to\linewidth{\color{gray!65}\fontsize{5}{5}\selectfont\leaders\hbox to3pt{\hss.\hss}\hfill}\par
  \vspace{2pt}\nointerlineskip
  \begingroup
  \renewcommand{\stackboxed}[3]{\begin{minipage}[c]{\stackcol}\makebox[\linewidth][c]{\fcolorbox{#3}{white}{\includegraphics[width=\dimexpr\linewidth-2\fboxrule\relax,trim=0bp 3bp 0bp 3bp,clip]{figures/hunyuan753_seed1/#2.jpg}}}\end{minipage}}
  \noindent\stackboxed{hyv}{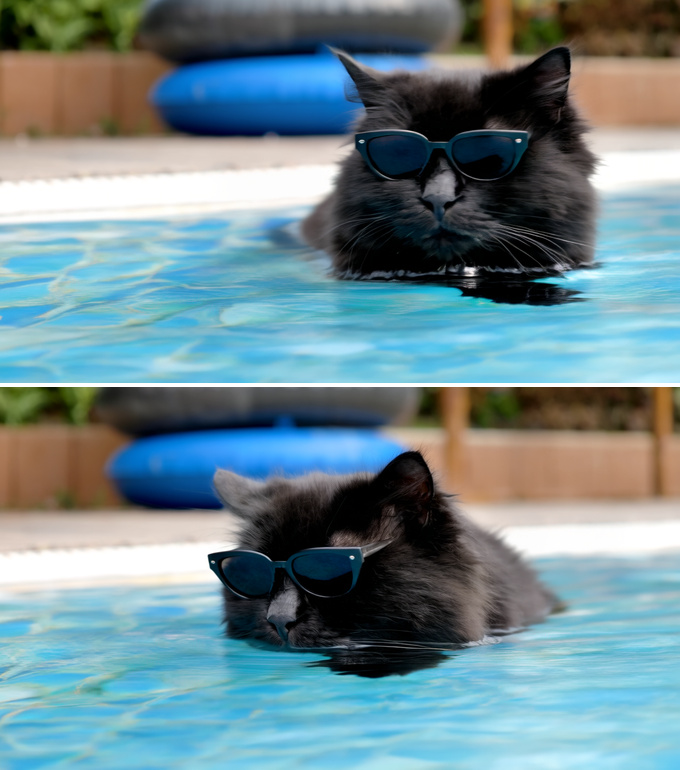}{densered}\hfill\stackboxed{hyv}{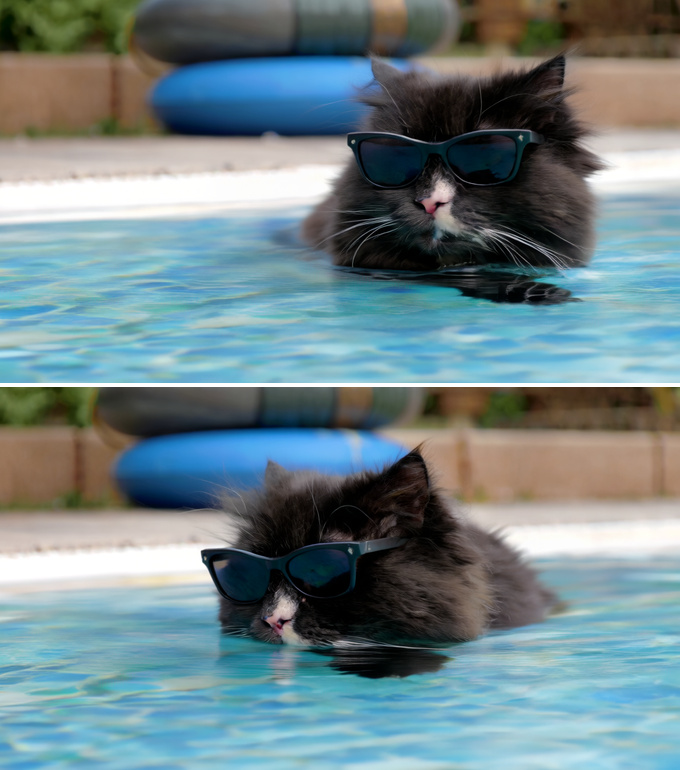}{boxgray}\hfill\stackboxed{hyv}{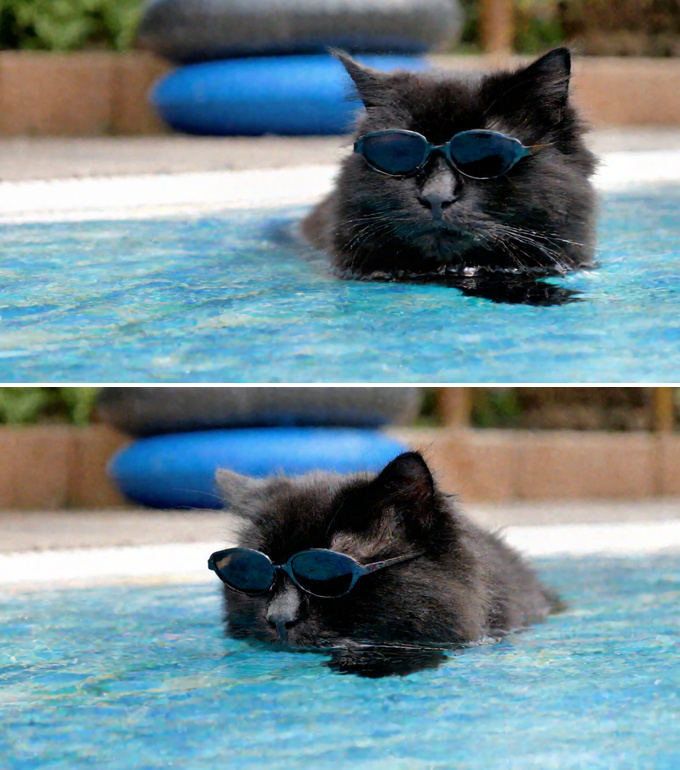}{boxgray}\hfill\stackboxed{hyv}{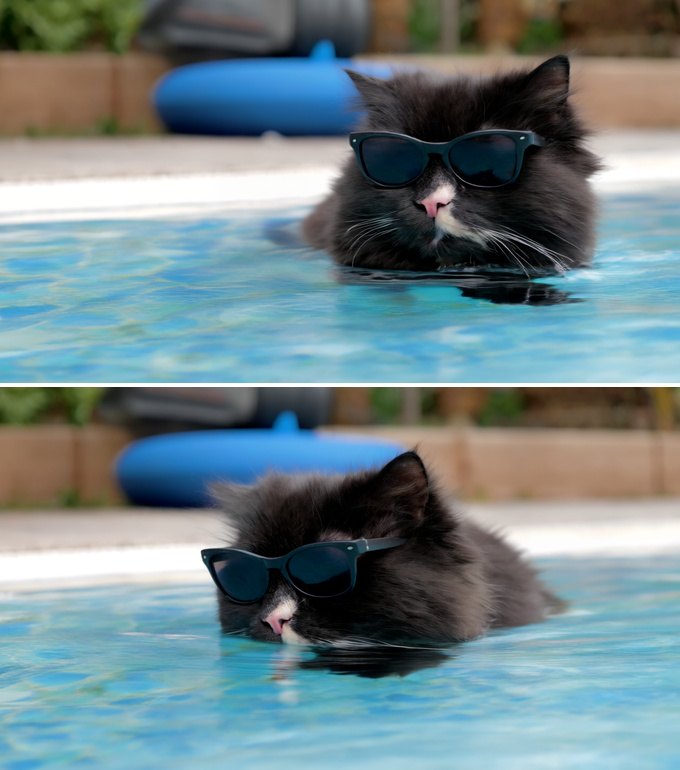}{boxgray}\hfill\stackboxed{hyv}{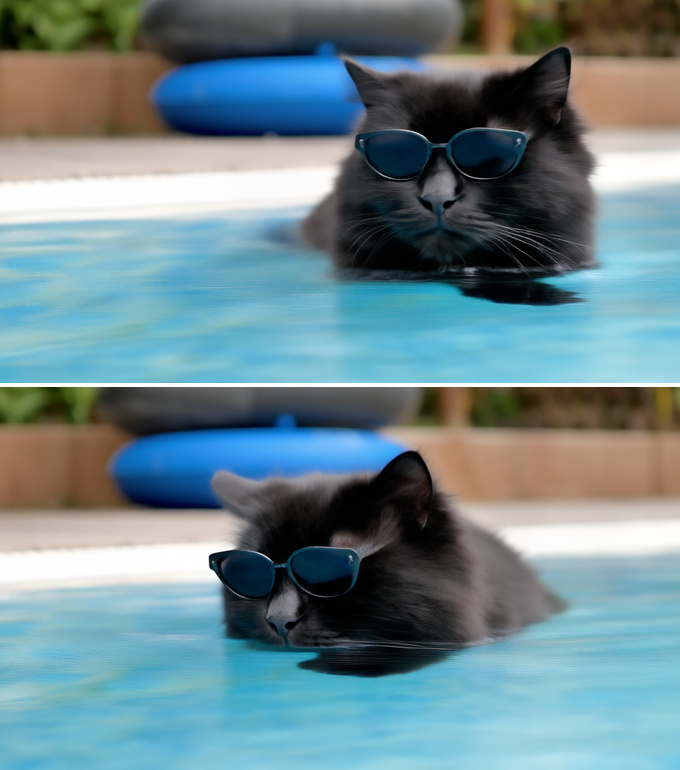}{boxgray}\hfill\stackboxed{hyv}{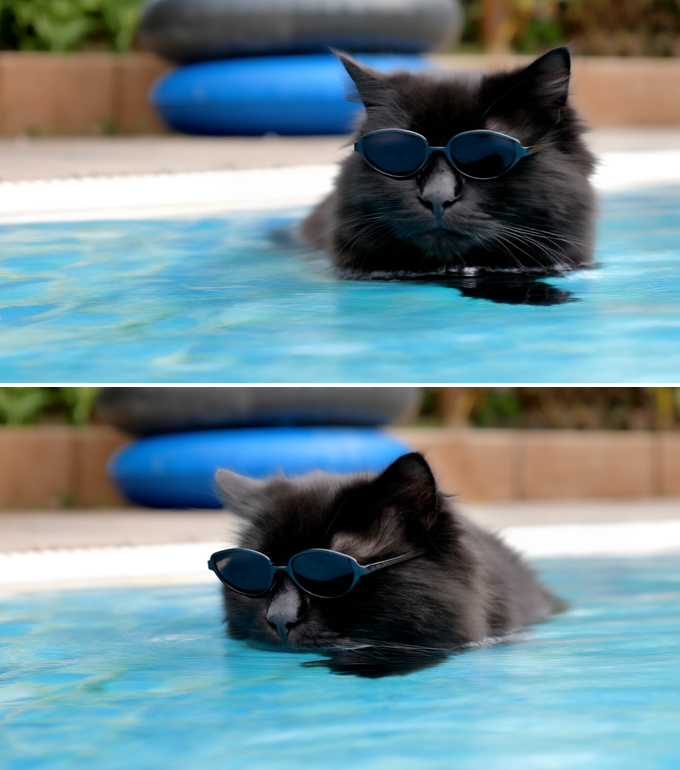}{blue!75!black}\par
  \endgroup
  \vspace{0.7pt}
  \noindent
  \stackdensemetric\hfill%
  \stackmetric{2.85}{21.88}{black}\hfill%
  \stackmetric{2.51}{26.78}{black}\hfill%
  \stackmetric{2.37}{22.15}{black}\hfill%
  \stackmetric{2.55}{26.65}{black}\hfill%
  \stackmetric{3.09}{29.25}{blue!75!black}\par
  \vspace{1pt}
  \stackfitprompt{A cat wearing sunglasses at a pool. Realistic, Natural lighting, Casual}

  \vspace{2pt}
  \nointerlineskip\hbox to\linewidth{\color{gray!65}\fontsize{5}{5}\selectfont\leaders\hbox to3pt{\hss.\hss}\hfill}\par
  \vspace{2pt}\nointerlineskip
  \begingroup
  \renewcommand{\stackboxed}[3]{\begin{minipage}[c]{\stackcol}\makebox[\linewidth][c]{\fcolorbox{#3}{white}{\includegraphics[width=\dimexpr\linewidth-2\fboxrule\relax]{figures/flux_aquarium_six/#2.png}}}\end{minipage}}
  \noindent
  \stacklabel{\textbf{Dense}}{densered}\hfill%
  \stacklabel{TeaCache}{black}\hfill%
  \stacklabel{NaviCache}{black}\hfill%
  \stacklabel{LeMiCa}{black}\hfill%
  \stacklabel{Spectrum}{black}\hfill%
  \stacklabel{\textbf{\method{}}}{blue!75!black}\par\vspace{2pt}
  \noindent\stackboxed{flux}{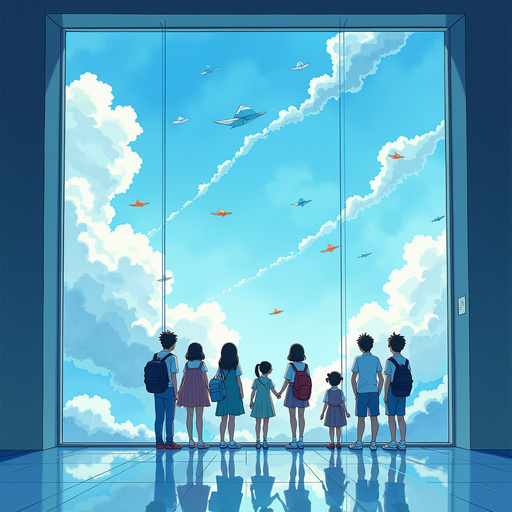}{densered}\hfill\stackboxed{flux}{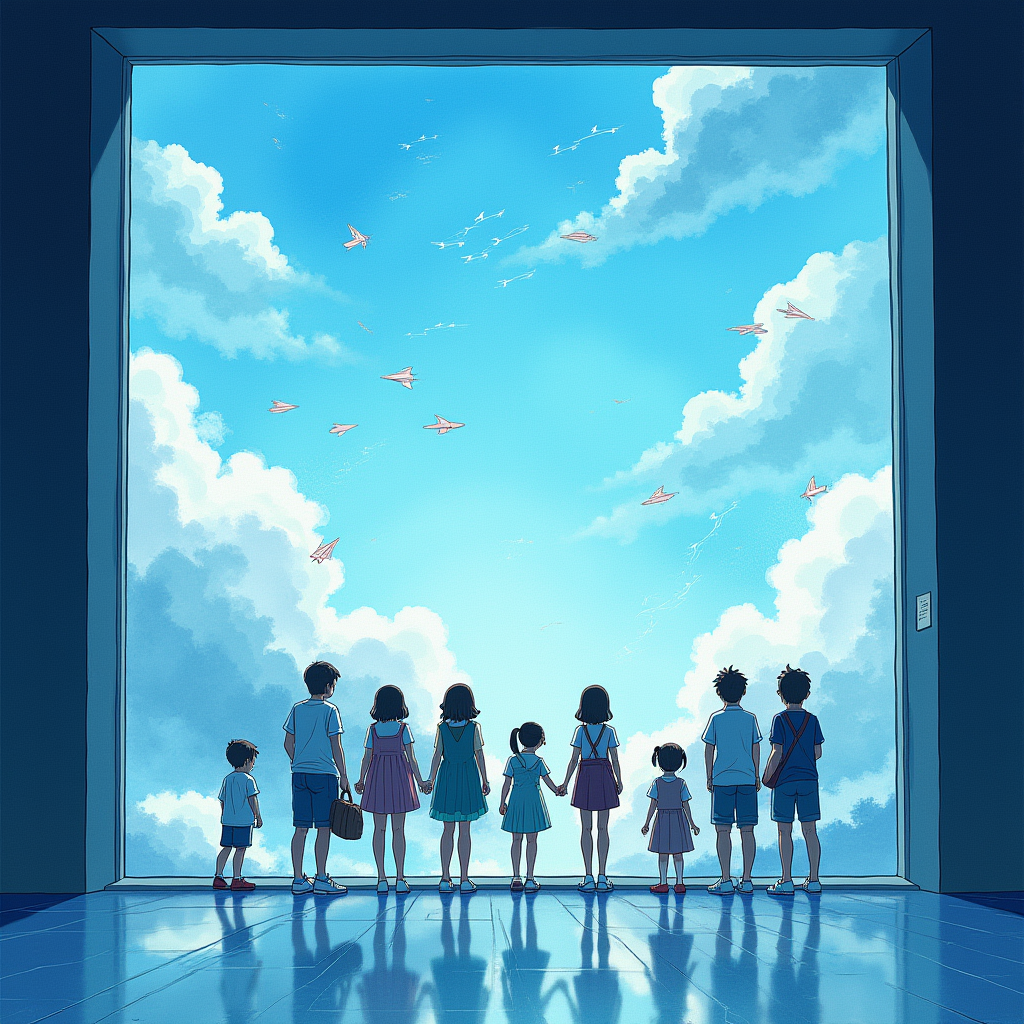}{boxgray}\hfill\stackboxed{flux}{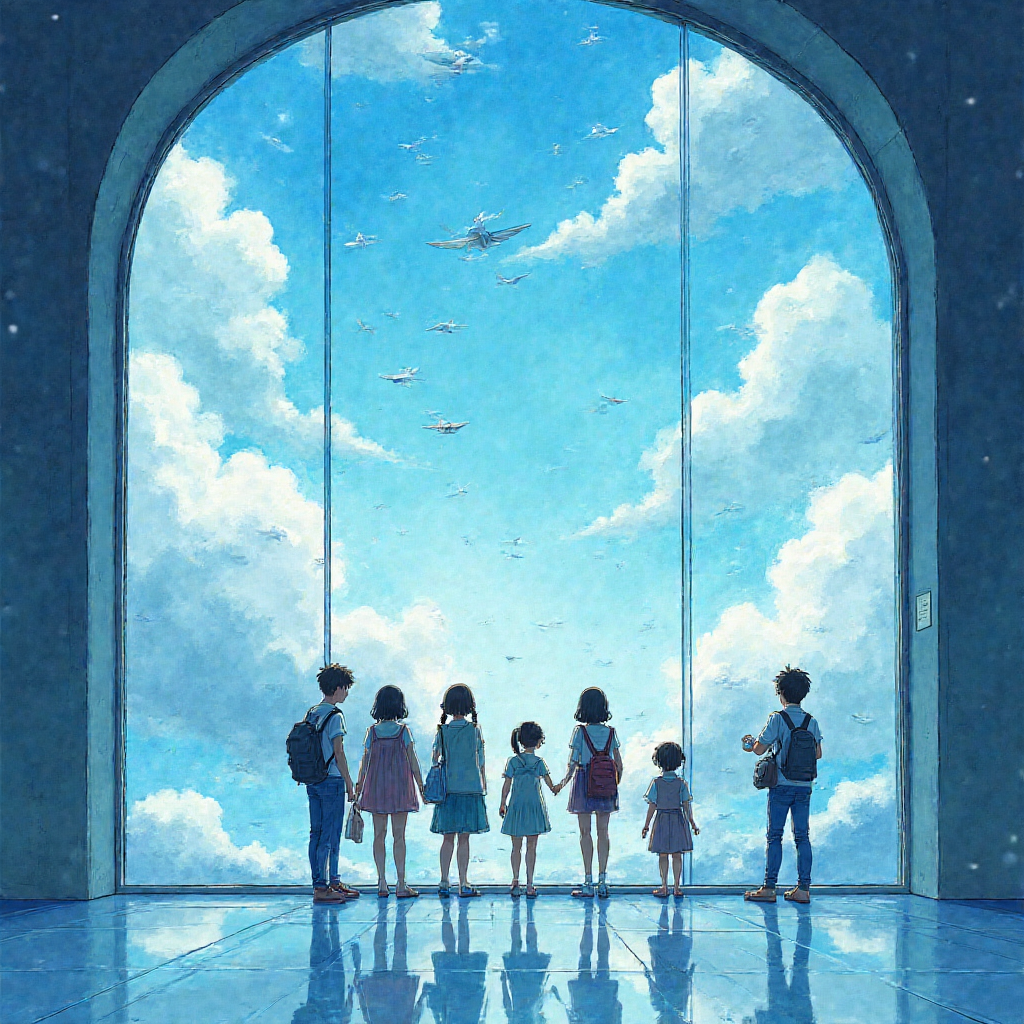}{boxgray}\hfill\stackboxed{flux}{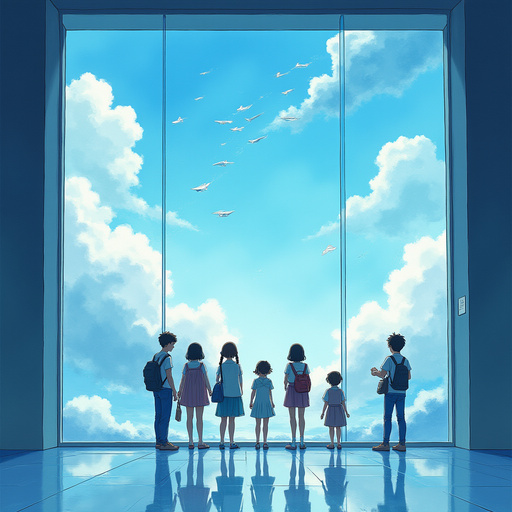}{boxgray}\hfill\stackboxed{flux}{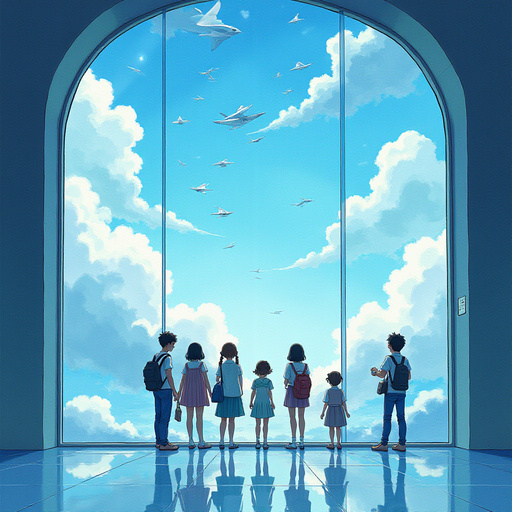}{boxgray}\hfill\stackboxed{flux}{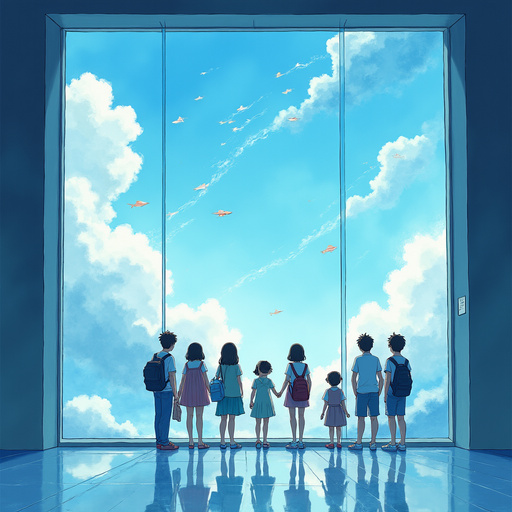}{blue!75!black}\par
  \vspace{0.7pt}
  \noindent
  \stackdensemetric\hfill%
  \stackmetric{3.15}{18.52}{black}\hfill%
  \stackmetric{3.21}{18.13}{black}\hfill%
  \stackmetric{3.27}{20.37}{black}\hfill%
  \stackmetric{3.38}{20.95}{black}\hfill%
  \stackmetric{3.42}{24.44}{blue!75!black}\par
  \vspace{1pt}
  \stackfitprompt{Dreamlike watercolor manga. Visitors stand before an enormous aquarium window, but clouds, miniature thunderstorms and flocks of paper airplanes drift behind the glass instead of fish. Blue sky light washes over the visitors and the polished floor. Complete environmental scene with character interaction, expressive visual storytelling, carefully drawn details, no text, no speech bubbles, no watermark.}
  \endgroup

  \endgroup
  \setlength{\abovecaptionskip}{4pt}%
  \captionof{figure}{Samples from CogVideoX-1.5 (top), HunyuanVideo (middle), and FLUX.1-dev (bottom). More image generation results are provided in Appendix Sec.~D. Best viewed zoomed in.}
  \label{fig:main-composition-results}
  \end{minipage}
\end{table}

\setcounter{topnumber}{2}
\section{Experiments}
\label{sec:experiments}
\vspace{-4pt}
\subsection{Experimental Setup}
\label{sec:experimental-setup}
\textbf{Models and baselines.}
We evaluate \method{} on CogVideoX-1.5-5B~\citep{yang2025cogvideox},
HunyuanVideo-13B~\citep{kong2024hunyuanvideo}, and
Wan2.1-1.3B~\citep{wan2025wan}, each with 50 steps. Wan2.1 results
are in Appendix Sec.~E.1. Baselines include
SVG~\citep{xi2025sparse}, Re-ttention~\citep{chen2025rettention},
SVOO~\citep{luo2026attention}, and DFSAttn~\citep{hu2026dfsattn} for sparse
attention and TeaCache~\citep{liu2025teacache},
LeMiCa~\citep{gao2025lemica}, and NaviCache~\citep{lv2026navicache} for caching. Pairing SVOO and DFSAttn with each caching baseline
yields six direct compositions. Appendix Sec.~D extends \method{} to FLUX.1-dev~\citep{blackforestlabs2024flux} image generation.
For image generation, we additionally compare with the caching method Spectrum~\citep{han2026spectrum}.

\textbf{Evaluation metrics.}
We use optimized VBench prompts~\citep{huang2024vbench},
with initial noise, guidance, and sampler settings matched across
methods within each comparison. Fidelity to dense generation is
measured using PSNR~\citep{hore2010image}, SSIM~\citep{wang2004image},
and LPIPS~\citep{zhang2018unreasonable} on paired lossless frames.
CLIP-IQA+~\citep{wang2023exploring} and VBench~\citep{huang2024vbench} assess generation
quality. For VBench, we average six quality dimensions after its official
per-dimension normalization, as detailed in Appendix Sec.~E.2.
The main tables also report the individual imaging quality (IQ), aesthetic quality (AQ), and
temporal flickering (TF) scores.

\textbf{Implementation details.}
The main experiments on CogVideoX-1.5 and Wan2.1-1.3B are conducted
on NVIDIA RTX 4090 GPUs, while those on HunyuanVideo use NVIDIA
A800 GPUs.
We use FlashAttention-2~\citep{dao2024flashattention2} for dense refreshes,
the Block Sparse Attention library~\citep{guo2024blocksparse} for
selected interactions, and custom Triton kernels~\citep{tillet2019triton}
for coarse-tail estimation and refinement.
\method{} requires no training or model fine-tuning. We select its
computation budget and error tolerances using the lightweight parameter
selection procedure in Appendix Sec.~C.4.
Appendix Sec.~C provides the full parameter settings for
each model and describes the memory optimizations.

\subsection{Comparison with Existing Methods}

\textbf{Quantitative results.}
\Tabref{tab:main-results-single} compares individual methods in the
normal-speedup regime. \methodQ{} improves PSNR over the
strongest individual baseline by 3.74\,dB on CogVideoX-1.5 and
1.42\,dB on HunyuanVideo. \methodS{} achieves the lowest latency
in this comparison and reaches speedups of 1.75$\times$ and
2.14$\times$ over dense generation, respectively.
Both settings outperform all individual baselines in fidelity
and VBench Overall on both models.
In the high-speed regime, \tabref{tab:lower-band} compares direct
sparse--cache compositions. \method{} exceeds the highest-PSNR
composition, DFSAttn with LeMiCa, by 3.78\,dB and 1.18\,dB on
the two models, respectively. It also achieves the best SSIM,
LPIPS, and VBench Overall scores alongside the lowest latency,
with speedups of 2.67$\times$ and 3.05$\times$. On CogVideoX-1.5,
\method{} achieves the best result on every reported metric.

\textbf{Qualitative results.}
The samples in \figref{fig:main-composition-results} show that
\method{} agrees more closely with dense generation in both fine details
and overall appearance. On CogVideoX-1.5, \method{} better preserves the
eagle's posture, the details of the snow-covered forest, and the overall
color of the scene, whereas three of the four compositions shift it toward a saturated
blue. On HunyuanVideo, it more faithfully reproduces the
details of the cat's face and the water, and renders the fur texture more clearly.
On FLUX.1-dev, it is the only method that preserves both the rectangular window and
the eight children. Additional comparisons across models and speedup regimes are provided in
Appendix Sec.~E.3.

\begingroup\renewcommand{\tabref}[1]{Tab.~10}% appendix table, unlinked
\begin{table}[t]
\caption{Ablation studies in the normal-speedup regime, with fast-regime results in
\tabref{tab:component-ablation-fast}. Latency is measured on one NVIDIA A800 GPU. Bold marks the best value within each comparison.}
\label{tab:component-ablation}
\centering
\captionsetup[subtable]{font=small,skip=3pt,position=bottom}
\setlength{\tabcolsep}{2pt}
\renewcommand{\arraystretch}{1.10}
\fontsize{8.5}{10}\selectfont
% Group-label rows sit slightly tighter and the right column slightly looser, so both columns
% reach about the same height; \vfill between (b) and (c) takes the small remainder.
\def\ablLabelGap{-2pt}\def\ablRightStretch{1.28}
% The right column is given the left column's height; \vfill absorbs the difference.
\setbox0=\hbox{%
\begin{minipage}[t]{.49\linewidth}
\vspace{0pt}
\begin{subtable}[t]{\linewidth}
\centering
\begin{tabularx}{\linewidth}{>{\raggedright\arraybackslash}Xrrr}
\toprule
Variant & PSNR$\uparrow$ & SSIM$\uparrow$ & LPIPS$\downarrow$ \\
\midrule
W/o refinement & 34.57 & 0.9749 & 0.0303 \\
Frozen state & 30.12 & 0.9396 & 0.0747 \\
\rowcolor{oursrow}
\method{} & \textbf{36.15} & \textbf{0.9817} & \textbf{0.0219} \\
\bottomrule
\end{tabularx}
\caption{Reconstruction components.}
\label{tab:ablation-components}
\end{subtable}
\par\vspace{4pt}
\begin{subtable}[t]{\linewidth}
\centering
\setlength{\tabcolsep}{1.2pt}%
\fontsize{7.5}{9}\selectfont
\begin{tabularx}{\linewidth}{>{\raggedright\arraybackslash}Xrrrrrr}
\toprule
& \multicolumn{3}{c}{\itshape Same locations} & \multicolumn{3}{c}{\itshape Matched runtime} \\
\cmidrule(lr){2-4}\cmidrule(l){5-7}
Variant & PSNR$\uparrow$ & SSIM$\uparrow$ & LPIPS$\downarrow$ & PSNR$\uparrow$ & SSIM$\uparrow$ & LPIPS$\downarrow$ \\
\midrule
PISA & 29.55 & 0.9351 & 0.0755 & 31.84 & 0.9549 & 0.0532 \\
Re-ttention & 26.93 & 0.8677 & 0.1329 & 28.68 & 0.8983 & 0.1036 \\
\rowcolor{oursrow}
\method{} & \textbf{36.15} & \textbf{0.9817} & \textbf{0.0219} & \textbf{36.15} & \textbf{0.9817} & \textbf{0.0219} \\
\bottomrule
\end{tabularx}
\caption{Alternative reconstructions.}
\label{tab:ablation-alternatives}
\end{subtable}
\end{minipage}%
}
\setbox2=\hbox{%
\begin{minipage}[t][\dimexpr\ht0+\dp0\relax][t]{.49\linewidth}
\vspace{0pt}
\renewcommand{\arraystretch}{\ablRightStretch}%
\begin{subtable}[t]{\linewidth}
\centering
\begin{tabularx}{\linewidth}{>{\raggedright\arraybackslash}Xrrr}
\toprule
Variant & PSNR$\uparrow$ & SSIM$\uparrow$ & LPIPS$\downarrow$ \\
\midrule
Dense-only history renewal & 31.80 & 0.9623 & 0.0468 \\
Uniform schedule & 35.70 & 0.9796 & 0.0254 \\
\rowcolor{oursrow}
\method{} & \textbf{36.15} & \textbf{0.9817} & \textbf{0.0219} \\
\bottomrule
\end{tabularx}
\caption{Prediction-history renewal and online scheduling.}
\label{tab:ablation-renewal}
\end{subtable}
\par\vfill
\begin{subtable}[t]{\linewidth}
\centering
\setlength{\tabcolsep}{0.5pt}
\begin{tabularx}{\linewidth}{>{\raggedright\arraybackslash}Xrrrr}
\toprule
Variant & PSNR$\uparrow$ & SSIM$\uparrow$ & LPIPS$\downarrow$ & Latency \\
\midrule
W/o prediction reuse & 38.79 & 0.9878 & 0.0143 & 845.51s \\
\mbox{\method{} {\fontsize{6}{7}\selectfont(matched latency)}} & \textbf{41.45} & \textbf{0.9934} & \textbf{0.0064} & 843.97s \\
\rowcolor{oursrow}
\method{} & 36.15 & 0.9817 & 0.0219 & \textbf{619.69s} \\
\bottomrule
\end{tabularx}
\caption{Prediction reuse.}
\label{tab:ablation-reuse}
\end{subtable}
\end{minipage}%
}
\noindent\usebox{0}\hfill\usebox{2}
\end{table}

\endgroup

\subsection{Ablation Studies}
\label{sec:ablations}

We ablate each component of \method{} on CogVideoX-1.5, namely attention
reconstruction, prediction reuse, and online scheduling. Each variant changes one
component with all other settings fixed. Appendix Sec.~E.4 repeats
every ablation at the fast setting of \tabref{tab:lower-band}, with the same conclusions.

\textbf{Attention reconstruction.}
The reconstruction in \secref{sec:reconstruction} pairs fresh coarse estimates
with cached dense refinement. Removing the refinement at the same computation
locations lowers PSNR from 36.15 to 34.57\,dB in \tabref{tab:ablation-components}.
Freezing the initial refinement and partition lowers it further to 30.12\,dB.
The refinement thus supplies detail missing from the coarse estimate and must be
renewed during sampling. Adapted PISA~\citep{li2026pisa} and
Re-ttention~\citep{chen2025rettention} change the cost of each sparse evaluation.
\Tabref{tab:ablation-alternatives} therefore compares them with our reconstruction
both at the same locations and at matched runtime. \method{} leads in both,
by 4.31 and 7.47\,dB at matched runtime.

\textbf{Prediction reuse.}
\Secref{sec:prediction-reuse} extrapolates reused steps from a prediction history
that corrected sparse evaluations also renew. Restricting this renewal to dense
refreshes lowers PSNR from 36.15 to 31.80\,dB in \tabref{tab:ablation-renewal},
showing that corrected sparse evaluations are valuable reuse sources. Disabling reuse and
computing all 50 steps raises PSNR to 38.79\,dB, but latency grows from
619.69 to 845.51\,s, so reuse saves 27\% of runtime. At matched latency, \method{}
instead reaches 41.45\,dB in \tabref{tab:ablation-reuse}.

\textbf{Online scheduling.}
The scheduler in \secref{sec:scheduling} places computed steps by prediction
error, concentrating them where predictions change quickly.
Spacing the same number of steps evenly lowers PSNR from 36.15 to
35.70\,dB in \tabref{tab:ablation-renewal}. Error-guided placement thus improves
fidelity at no extra cost.

\section{Conclusion}
\label{sec:conclusion}
We presented \method{}, a training-free method that couples sparse
attention and hierarchical caching through reciprocal correction instead of composing them directly. We
showed that the accuracy of cache sources affects the fidelity of the
composition and that sparsification errors persist across nearby steps.
Building on these findings, \method{} maintains a two-level cache
hierarchy. At the attention level, cached dense refinement corrects sparse
attention, while fresh sparse statistics regulate how strongly this aging
refinement is applied. At the prediction level, corrected predictions
renew the sources for whole-step reuse. Online error feedback
decides when to compute and when to refresh densely, so the schedule
adapts to each sample without costly offline search. Across three video
diffusion models, \method{} improves the speed--fidelity trade-off over
individual accelerators and direct compositions, reaching a $3.05\times$
speedup on HunyuanVideo with higher fidelity than all tested compositions.

\newpage
\textfloatsep=\recapOrigTextfloatsep
\let\section\recapOrigSection
\let\subsection\recapOrigSubsection

\section*{AI use statement}

In this work, we used generative AI tools for polishing language to improve clarity. We have not used generative AI tools to generate research ideas, conduct the reported
experiments, or generate experimental results. We have reviewed all
AI-assisted work, checking revisions for technical accuracy and consistency
with our method and experimental evidence. We take responsibility for the
final content of this work, including text, claims, or artifacts produced
with the aid of generative AI.

\section*{Ethics statement}

This research uses publicly available
video-generation models and focuses on algorithmic improvements for inference
acceleration. We identify no ethical concerns specific to the proposed
acceleration method.

\section*{Reproducibility statement}

Method details are provided in \secref{sec:method},
Appendix Sec.~A, and Appendix Sec.~B.
Experimental configurations and evaluation metrics are described in
\secref{sec:experimental-setup} and Appendix Sec.~C.
Code and video samples will be available at \url{https://github.com/lyraxtg/RECAP}.

\bibliographystyle{iclr2027_conference}
\bibliography{references}

\end{document}